\documentclass[runningheads,a4paper]{llncs}

\usepackage{amssymb}
\usepackage{graphicx}

\usepackage{url}
\urldef{\mailsa}\path|{alfred.hofmann, ursula.barth, ingrid.haas, frank.holzwarth,|
\urldef{\mailsb}\path|anna.kramer, leonie.kunz, christine.reiss, nicole.sator,|
\urldef{\mailsc}\path|erika.siebert-cole, peter.strasser, lncs}@springer.com|

\begin{document}

\mainmatter  

\title{Sparse Over-complete Patch Matching}

\titlerunning{Akila Pemasiri}

%
%

\author{Akila Pemasiri\inst{1} \and
Kien Nguyen \inst{1} \and \\
Sridha  Sridharan\inst{1}  \and Clinton Fookes\inst{1}}
\authorrunning{Akila Pemasiri et al.}
%

\institute{Image and Video Research Lab, Queensland University of Technology, Brisbane, Australia\\
\email{\{a.thondilege,k.nguyenthanh,s.sridharan,c.fookes\}@qut.edu.au}\\
}
%

%
%

\toctitle{Lecture Notes in Computer Science}
\tocauthor{Authors' Instructions}
\maketitle

\begin{abstract}
Image patch matching, which is the process of identifying corresponding patches across images, has been used as a subroutine for many computer vision and image processing tasks. State -of-the-art patch matching techniques take image patches as input to a convolutional neural network to extract the patch features and evaluate their similarity.  Our aim in this paper  is to improve on the state of the art patch matching techniques by observing the fact that a sparse-overcomplete representation of an image posses statistical properties of natural visual scenes which can be exploited for patch matching. We propose a new paradigm which encodes image patch details by encoding the patch and subsequently using  this sparse representation as input to a neural network to compare the patches. As sparse coding is based on a generative model of natural image patches, it can represent the patch in terms of the fundamental visual components from which it has been composed of, leading to similar sparse codes for patches which  are built from similar components. Once the sparse coded features are extracted, we employ a fully-connected neural network, which captures the non-linear relationships between features, for comparison. We have evaluated our approach using the \textit{Liberty} and \textit{Notredame} subsets of the popular UBC patch dataset and set a new benchmark outperforming all state-of-the-art patch matching techniques for these datasets. \footnote{This paper is under consideration at Pattern Recognition Letters.} 

\end{abstract}

\section{Introduction}
\label{introduction}
 Image patch matching is a fundamental process used in many  image processing applications such as image classification \cite{yao2012codebook}, object recognition \cite{keysers2007optimal}   image stitching \cite{brown2007automatic},  and correspondence estimation required for structure from motion \cite{1}. Depending on the application, a patch matching technique may have to deal with different challenges such as  intra class variations, different lighting conditions, shadings and occlusions \cite{nowak2007learning}. Patch matching techniques have evolved through decades while trying to tackle these challenges \cite{brown2011discriminative,trzcinski2012learning,simonyan2014learning}.
 
In the process of patch matching,  two key components can be identified: the patch descriptor and the matching function. The patch descriptor is extracted to represent the features of the  image patch which are then  subjected to the matching function to estimate their similarity. The patch matching techniques have evolved from early stages across both of these components. In the early stages     the pixel values were used as the patch features and the  $L_{1}$ distance among pixel values  were used to measure the similarity. These types of early stage approaches have   been heavily sensitive to slight transformations of the patches  \cite{shakhnarovich2005learning}. The  state-of-the-art patch matching techniques are based on  deep learning  where the    patch descriptor identification and similarity estimation is performed using deep neural networks \cite{han2015matchnet,zagoruyko2015learning}. These techniques aim to tackle the challenging nature of patch matching including  viewpoint change, illuminations variations and different shadings. However the state-of-the-art patch matching techniques predominately focus on  visual elements that are present in the current image and they do not effectively consider  the  composition of underlying object shapes and structures that have contributed to the current image. 

Sparse coding is an important concept that is used for the sensory coding in the human brain \cite{field1994goal}. In human visual cortex, for each information captured by human eye a  neural code is formed by activating a selected set of neurons from a large neural population.
Sparse coding has revealed how the human brain can process a new image through the relations it can determine with images it has already come across so far. In this paper, we draw inspiration from this to propose a mechanism where sparse coding is utilised to produce a patch descriptor. In contrast to the state-of-the-art patch matching techniques which use visual deep features of the patch image \cite{han2015matchnet,zagoruyko2015learning}, we consider the statistical representation of the patch image in deducing the descriptor. The study carried out by Gregory Shakhnarovich \cite{shakhnarovich2005learning} has concluded that in overcomplete sparse coding   a small variation in an image can cause a significant change in the resulting coefficients. Therefore a hand engineered similarity measure may be inadequate for comparing  the resultant  sparse codes.  To accommodate for these nonlinear relationships we use a neural network based mechanism to determine the patch similarity as a neural network is capable of dealing with the heavy variations in the input \cite{tu1996advantages}.

The remainder of the paper is organized as follows. Section \ref{lit} discusses the most recent approaches used for patch matching. Section \ref{methodology} elaborates the methodology that we propose and  Section  \ref{experiments} describes the experiments carried out to compare our method. In Section \ref{conclusion} we conclude the paper.

\section{Related work}
\label{lit}

The previous work related to image patch matching can be categorized  under three categories: approaches  using image intensities, approaches using hand engineered features and  approaches using deep learned features. \textbf{Traditional patch matching techniques} start with using pixel based distance to identify patch correspondence, where $L_{1}$ distance was used to compare pixel values of two  patch images. The next stage of this category image patch matching has used normalized correlation between patches to identify their correspondence \cite{nakhmani2013new,bart2004class}. Calculating image descriptors such as Scale-invariant feature transform (SIFT) \cite{lowe2004distinctive} and DAISY \cite{tola2008fast} and estimating the descriptor distance is the main concept behind the hand engineered feature based methods .

\begin{figure}[!ht]
\centering
   
   \includegraphics[width=1\linewidth]{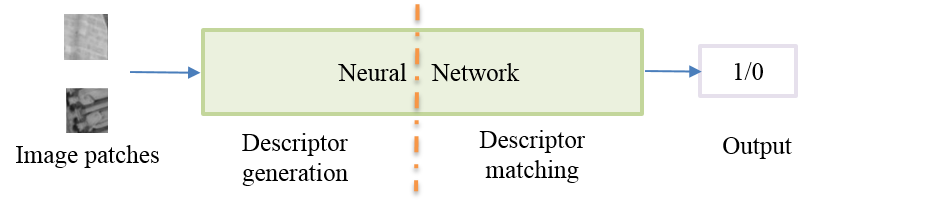}
   \caption{State-of-the-art approach}
   \label{stoa} 
\end{figure}

\begin{figure}[!ht]
\centering
   
   \includegraphics[width=1\linewidth]{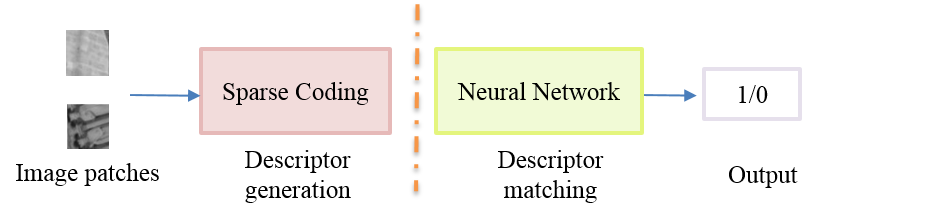}
   \caption{Our approach}
   \label{ourapproach} 
\end{figure}

In the most recent literature, there are two main approaches which have been motivated by the recent advances in neural networks and deep learning \cite{han2015matchnet,zagoruyko2015learning}. The aim of these \textbf{deep learning based patch matching techniques}  are to  generate more robust descriptors which can overcome the drawbacks of hand crafted features such that the descriptors and the matching algorithms are not vulnerable to  challenging factors in the patches such as illumination changes , occlusions and shadings.

In the approach suggested by Zagoruyko and Komodakis \cite{zagoruyko2015learning} they have evaluated three main neural network architectures for the patch matching. The architectures they have suggested are 1) Siamese, 2) Pseudo-siamese and 3) 2-channel. The siamese and pseudo-siamese architectures contain two branches and as the input each branch takes one of the two patches to be compared. The output of these  two branches are analogous to the feature descriptors in the traditional approaches and the branches  are merged at the top to make the comparison. In contrast to the siamese architecture  where the weights of the two branches are shared in the pseudo siamese architecture the weights are uncoupled. In 2-channel architecture  two patches are considered as a 2 channel image where there is no explicit separation on feature descriptor generation and matching.  The evaluations on these architectures have been carried out on UBC dataset \cite{brown2011discriminative}. Their evaluations  conclude that when the convolution layer was divided into small kernels of size $ 3\times 3$, 2 channel architecture performs the best.

Matchnet \cite{han2015matchnet} is a convolution neural network (CNN) based approach where the architecture consists of two sub components as feature network - to extract the features of the patches and metric network - to model the similarity between the patches. To train the networks they have used cross entropy error. Their evaluations are also based on the standard UBC dataset \cite{brown2011discriminative}.

Coefficients of an encoding mechanism for a set of images such that the accuracy of reconstruction   and the sparseness of coefficient are maximized, can possess statistical properties of natural visual scenes \cite{olshausen1997sparse,hyvarinen2001two}. \textbf{Sparse coded coefficients} have successfully been used in computer vision applications as well as in signal processing applications in general. While some applications of this concept  are image classification \cite{yang2009linear}, image reconstruction \cite{shang2008image} and audio analysis \cite{adiloglu2012graphical,scholler2011sparse}, this concept has not been adopted for patchmatching processes. In the previous approaches of image patch matching visual features have majorly  been used. In our approach  we use coefficient resulted from over-complete sparse coding of patches (Figure \ref{stoa}, Figure \ref{ourapproach}) as the feature representation and to estimate the patch similarity we use a neural network while tolerating non linear dependencies in input images.

\section{The proposed patch matching framework}
\label{methodology}
\begin{figure}[!ht]
\begin{center}
   \includegraphics[width=\linewidth]{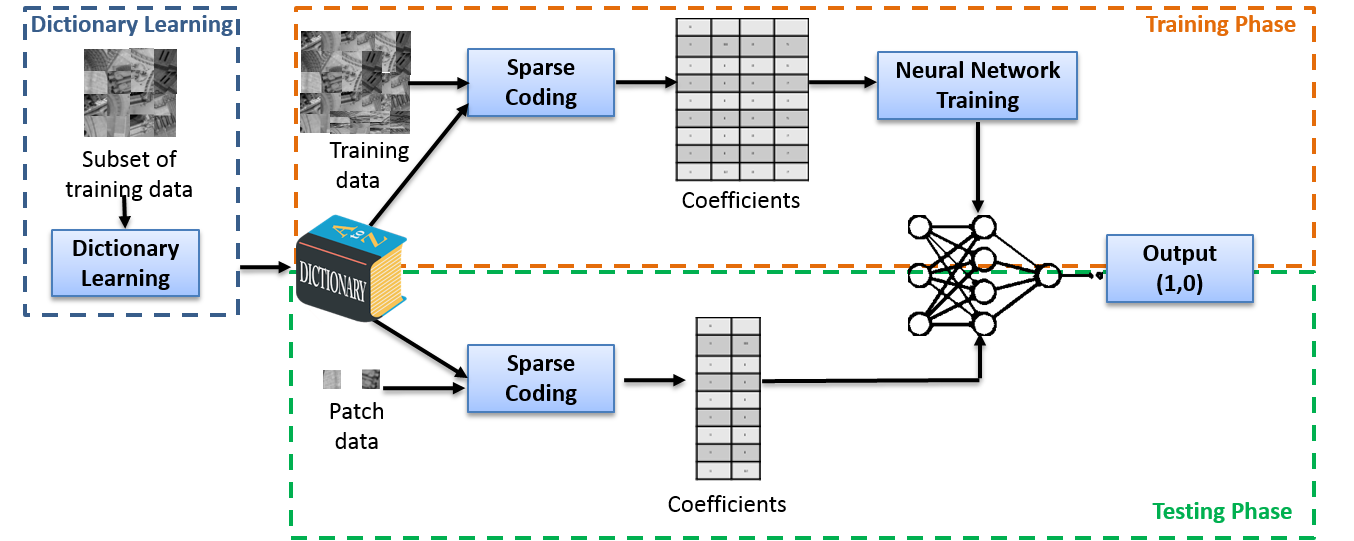}
\end{center}
   \caption{The architecture of patch matching neural network. First phase is the \textit{`Dictionary Learning'} where the set of bases are found. In the second phase \textit{`Training Phase'} encoded data(coefficients) using the learnt dictionary is used to train a neural network. In the \textit{`Testing Phase'} the encoded pairs of patches will be fed into the trained model where the network will output 1 or 0 depending on patch similarity. }
\label{workflow}
\end{figure}

The overall flow of our patch matching framework is depicted in Figure \ref{workflow}. First we learn a set of basis (dictionary) based on a randomly selected  sample of  training data. This dictionary is used to encode the training data as well as the test data. Once the training data has been encoded we use that data to train a neural network in a supervised manner. Then at the testing phase we encode the test data and feed them to the trained model to examine the patch similarity. Section \ref{descriptor} elaborates the encoding technique that we used including dictionary learning phase (Section \ref{basislearning}) and coefficient learning phase (Section \ref{findingc}). The details on  the  neural network architecture that we used for patch matching is elaborated in Section  \ref{patchmatch}. 

\subsection{Patch descriptor }
\label{descriptor}
As discussed in the related work section, our approach use  over complete sparse representation of image patch as the patch descriptor. In the sparse coding, the dictionary is also denoted as the basis vector and the representation of the image patch with respect to the basis is denoted as the coefficient vector. The term sparse enforces the number of non zero elements in the coefficient vector to be minimized while offering a good discriminative power \cite{ballard1993object,ren2005learning}. The problem of learning the basis vector and the coefficients can be formed as Equation \ref{sparse1} :

\begin{equation}
    X\approx BC,
    \label{sparse1}
\end{equation}

where $ X = \left [ x_{1}, ..., x_{n} \right ]\epsilon \, \mathbb{R}^{m\times n} $ is the data matrix, $ B$ is the basis vector such that $ B \, \, \epsilon \, \, \mathbb{R}^{m\times k}  $ and $C$ is the coefficient vector such that $ C \, \, \epsilon \, \, \mathbb{R}^{k\times n}  $. When $ m > k$ it is known as \textit{undercomplete} , $m = k$ \textit{complete} and $m < k$ \textit{overcomplete}. 

To ensure the minimized reconstruction error, the Equation \ref{sparseenergy}

\begin{equation}
\smash{\displaystyle\min_{B,C}} \left \|X -BC  \right \|,
\label{sparseenergy}
\end{equation}
has to be solved and to ensure the overcompleteness $k$ should be greater than $m$. In this work we employ `overcomplete' representation as it has been identified overcomplete representation is better able of capturing patterns that are present in the input image while ensuring a better robustness at the presence for factors such as image degradations, scale changes, translations and rotations \cite{olshausen1997sparse}.

To solve the sparse coding problem we used the method similar to the method  used by Cia \textit{et al.}
\cite{cai2011sparse}. The methods that were used to construct the basis vector and coefficient vector are discussed in detail in next two subsections, Subsection \ref{basislearning} and Subsection \ref{findingc} respectively.

\subsubsection{Learning the basis vector (Dictionary learning)}
\label{basislearning}

The objective of basis learning is to identify the underlying structure of the data. Many algorithms have been devised to obtain this dictionary \cite{belkin2001laplacian,tenenbaum2000global} and they have used nearest neighbour graphs to model the geometric structure of the images. In our approach to model the  underline structure we used a graph where the heat kernel function was used as the edge weights \cite{li2016adaptive}. For the graph construction we used the Equation

\begin{equation}
W\left( ij \right )=\left\{
                \begin{array}{ll}
                  exp^{( -\frac{\left \| x_{i} -x_{j} \right \|_{2}}{t} )},\, \,  \, x_{i} \, \epsilon \, \mathcal{N}_{p}(x_{j}) \, \,or \, \, x_{j} \, \epsilon \, \mathcal{N}_{p}(x_{i}) \\
                  0, \, \, \, \, \, \, \, \, \, \, \, \, \, \, \, \, \, \, \, \,  \, \, \, \, \, \,\, \, \, \, \, \, \, \,  \, \, \, \, \, \, otherwise\\
                  \end{array},
              \right.
    \label{graphconstruction}
\end{equation}
 where $\mathcal{N}_p(x)$ stands for the set of $p$ nearest neighbours of $x$. 

We defined a diagonal matrix D such that $D_{ii} = \sum _{j} W_{ij}$ and based on that we define the graph Laplacian \cite{chung1997spectral}  as

\begin{equation}
    L = D - W.
    \label{laplacian}    
\end{equation}
To find the a flat embedding of data points we formulate the generalized eigenvector problems as \cite{belkin2001laplacian},

\begin{equation}
    Ly = \lambda Dy.
    \label{generalizedeigenvector}
\end{equation}

Let $Y' = [y_{1}, y_{2},..........y_{p}]$ be the solutions of Equation \ref{generalizedeigenvector}, we consider  $Y = [y_{1} .....  y_{k}]$ a subset of eigenvectors denoted by $Y'$, such that $Y$ contains   the k eigen vectors with the largest magnitude eigen values among $Y'$. Following the definition by Belkin and Niyogi \cite{belkin2001laplacian} we consider this $Y$ as the flat embedding of image data. To obtain $B$ which fits the data best we solved the optimization function  

\begin{equation}
    \smash{\displaystyle\min_{B}} \left \|Y -X^{T}B  \right \|^2 +\alpha \left \| B \right \|^2,
    \label{sccfunction}
\end{equation}
where $\alpha$ is used as regularization parameter to reduce overfitting.

\subsubsection{Learning the coefficient vector}
\label{findingc}
Once the basis vector, $B$ is obtained, to find the coefficient vector $C$ we used the cost function

\begin{equation}
    \smash{\displaystyle\min_{c_{i}}}\left \|x_{i} -Bc_{i}  \right \| ^{2} +\beta \left | c_{i} \right |,
    \label{findingcoefficient}
\end{equation}
which can be interpreted using two terms as \textit{Reconstruction term} and \textit{Sparsity penalty term}. The aim of the  reconstruction term is to yield a good representation of data while minimizing the reconstruction error and the aim of the sparsity penalty term is to enforce the sparsity. In Equation \ref{findingcoefficient} $x_{i}$ refers to the image $i$ and $c_{i}$ refers to the coefficient vector corresponding to $x_{i}$.

To solve the optimization problem in Equation \ref{findingcoefficient} we used least angle regression algorithm \cite{efron2004least}.

\subsection{Patch matching network}
\label{patchmatch}
The features encoded through sparse coding  are used to train a neural network in the supervised manner by minimizing the binary cross entropy error (Equation \ref{binarycrossentropy}) using Adam optimizer \cite{kingma2014adam} which is an improved version of Stochastic gradient decent. In Equation \ref{binarycrossentropy}, $y_{i}$ which is the label for each patch pair can have the values 1/0, and $\hat{y_{i}} $ stands for the output of the neural network where $N$ is the number of patch pairs that was subjected to training.

\begin{equation}
    E = -\frac{1}{N} \sum_{n=1}^{N} \left [ y_{n}\,\,log \,\, \hat{y_{n}} + (1 - y_{n})\,\, log \,\, (1 - \hat{y_{n}}) \right ]
    \label{binarycrossentropy}
\end{equation}

To compare the patches we evaluated two neural network architectures while trading off between processing power, flexibility with complexity. The two architectures we evaluated referred as \textit{Architecture 1} and \textit{Architecture 2} here on wards where the first one contains three hidden layers with 500, 80 and 4 neurons  followed by a sigmoid neuron and the second one contains  one hidden layer with 1000 neurons followed by a sigmoid neuron. The use of neural network in the patch matching enables to model the nonlinear relations between the input and output with the used of hidden neurons.  Convolutional Neural Network(CNN) based architectures were not considered in our work due to the fact that they work with 2D inputs \cite{o2015introduction} where as  in this task our input is a 1D coefficient vector. In addition, Recurrent Neural Network(RNN) and Long Short Term Memory(LSTM) based architectures   were not considered as they are targeted for sequence predictions \cite{le2015tutorial} where as patch matching is not a sequential process. 

\section{Experiments}
\label{experiments}
\textbf{Dataset}: For the evaluation we used the standard UBC patch dataset \cite{brown2011discriminative} which contains of three subsets named as \textit{Liberty}, \textit{Notredame} and \textit{Yosemite}, which contain image patches captures at the Liberty statue, Notredame cathedral and Yosemite valley. Each of these subsets contained anotated patches so that there are 50\% of matching pairs. This dataset has been used as the bench mark for many patch matching frameworks \cite{han2015matchnet,zagoruyko2015learning,brown2011discriminative,jia2011heavy}. For the evaluations we used the first two subsets as the groundtruth annotation of those two subsets are publicly available. 
\newline

\noindent   \textbf{Evaluation protocol}: We used the commonly used evaluation protocol where training was carried out on a set patches retrieved from one subset of the dataset and the testing was carried out on a  set of patches retrieved from the other subset. To measure the accuracy the  false positive rate at 95\% true positive rate (TPR, Recall) which is also known as Error@95\% was used.
\newline

\noindent  \textbf{Experimental setting}: We learnt a dictionary for the sparse coding for \textit{Liberty} and \textit{Notredame} datasets separately. For the dictionary learning we randomly picked 50k patch pairs (100k patch images) on each subset and  the value of $k$ was changed to evaluate how the accuracy get affected by the change of the overcompleteness. We have evaluated our method for 3 values of $k$ as 4097, 5000, 6000 and 7000 where the original dimension was 4096. The resultant dictionaries were in the size of $4096 \times 4097$, $4096\times5000$, $4096\times6000$ and $4096\times7000$. For the dictionary encoding process regularization parameter was set to 0.1. 

\begin{figure}
\minipage{0.47\textwidth}
  \includegraphics[width=\linewidth]{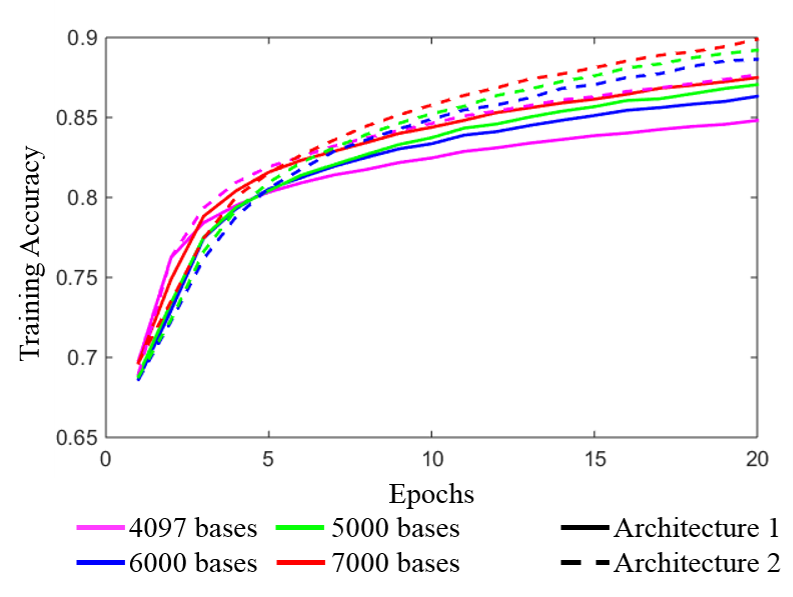}
  \caption{Training accuracy on \textit{Liberty}  dataset.}\label{libtrain}
\endminipage\hfill
\minipage{0.47\textwidth}
  \includegraphics[width=\linewidth]{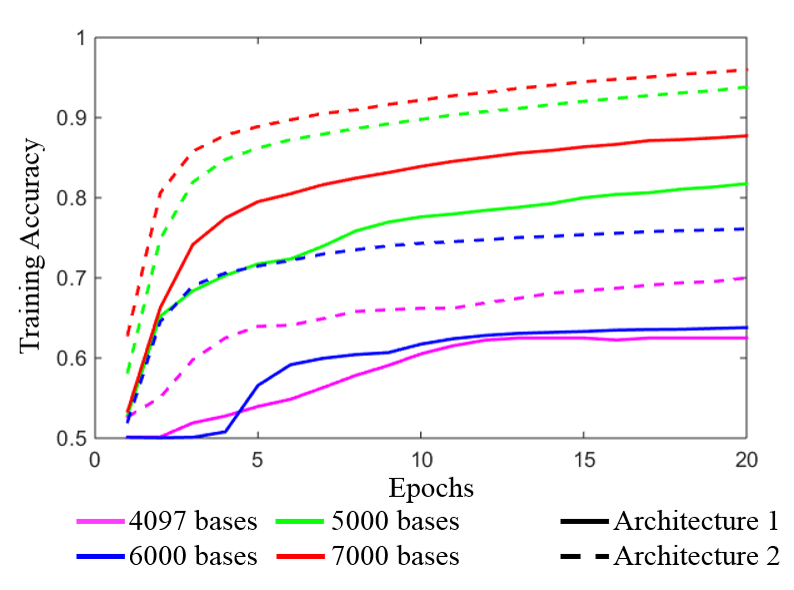}
  \caption{Training accuracy on \textit{Notredame} dataset.}\label{notretrain}
\endminipage
\end{figure}

\begin{figure}
\minipage{0.47\textwidth}
  \includegraphics[width=\linewidth]{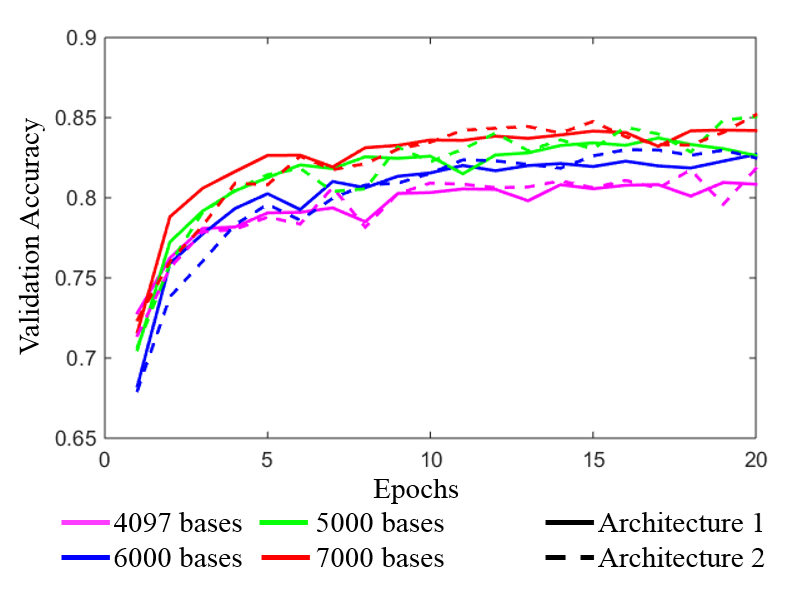}
  \caption{Validation accuracy on \textit{Liberty}  dataset.}\label{libvalidation}
\endminipage\hfill
\minipage{0.47\textwidth}
  \includegraphics[width=\linewidth]{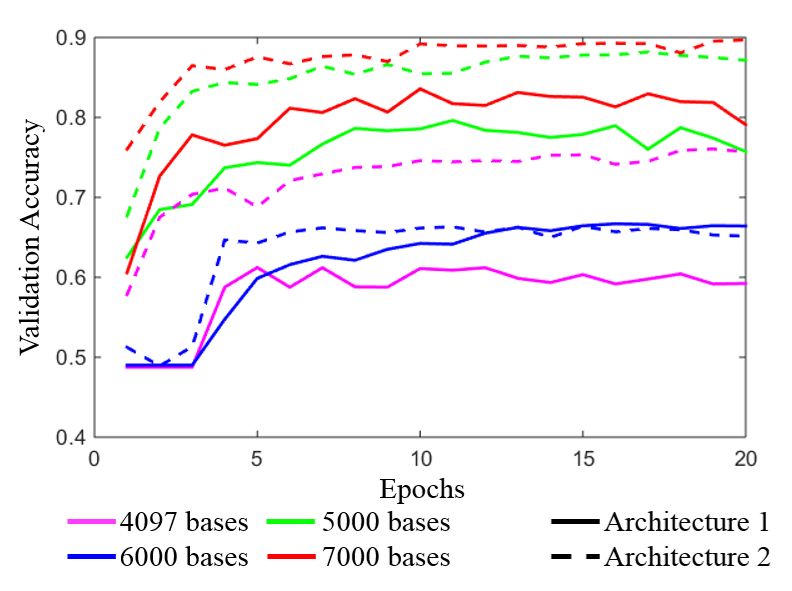}
  \caption{Validation accuracy on \textit{Notredame}  dataset.}\label{notrevalidation}
\endminipage
\end{figure}

For the training we use 500k patch pairs for each subset and encoded it using the created dictionary for that particular subset. Once the patch images are encoded, the corresponding patch images of each pair was concatenated. This was used for the training and validation of the neural network with a validation, training split ratio of 1:4, with a batch size of 64. The training accuracy  related to these settings are depicted in Figure \ref{libtrain} and Figure \ref{notretrain} while the and validation accuracy are depicted in Figure \ref{libvalidation} and Figure \ref{notrevalidation}.

\begin{figure}
\minipage{0.47\textwidth}
  \includegraphics[width=\linewidth]{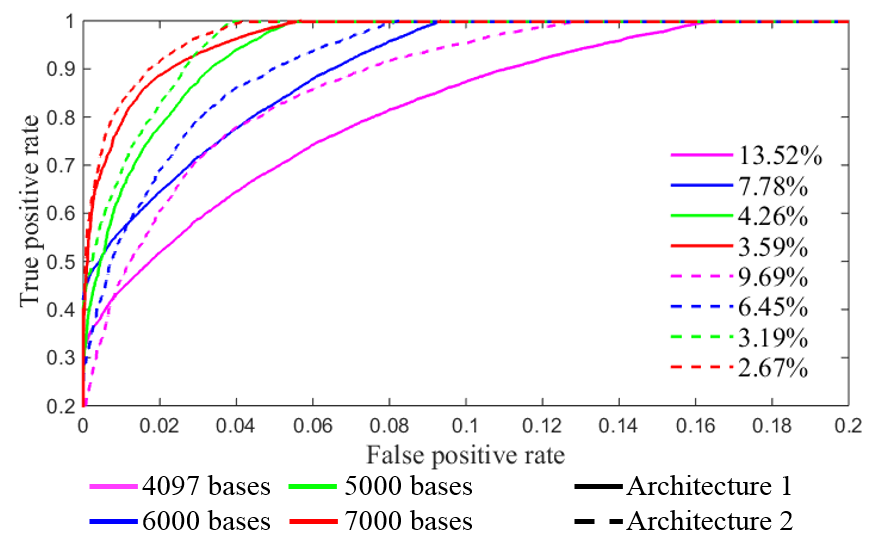}
  \caption{ROC curves on different experiment settings with their error@95\%  for \textit{Liberty}  dataset.}\label{rocliberty}
\endminipage\hfill
\minipage{0.47\textwidth}
  \includegraphics[width=\linewidth]{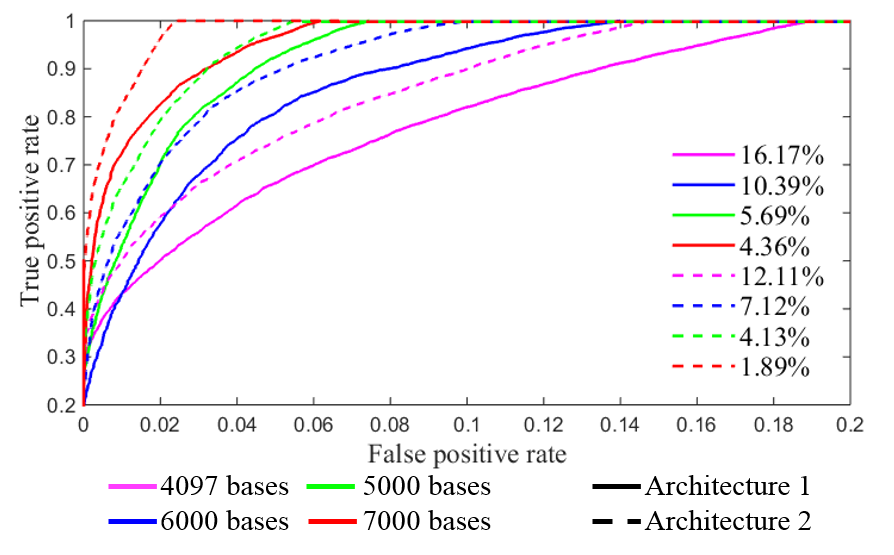}
  \caption{ROC curves on different experiment settings with their error@95\%  for \textit{Notredame}  dataset.}\label{rocnotre}
\endminipage
\end{figure}

From these observations it can be seen that an higher training accuracy a  has been obtained for the experiments where the number of basis has been set to 7000 for both subsets of data as well as both neural network architectures. The highest accuracy values obtained at the validation time is indicated in Table \ref{maxumumvalidation} and it can be seen that the recorded highest validation accuracy has been obtained for the experiments with 7000 bases. Furthermore when the results from two network architectures are analyzed it can be seen that the architecture 2 has yielded better training and validation accuracy values.

As the test data, we used 100k patch pairs from each of \textit{Liberty} and \textit{Notredame} datasets, which were not subjected to training. In order to evaluate our approach, we used the dictionary learnt for the \textit{Notredame} dataset to encode the \textit{Liberty} test dataset and used it as the input to the neural network trained for the \textit{Notredame} dataset. Similarly we tested the \textit{Notredame} test set using the \textit{Liberty} dictionary and the trained model. The Receiver operating characteristic (ROC) curve related to the experiments are illustrated in Figure \ref{rocnotre}  and Figure \ref{rocliberty} where the values of error@95\% with related to different overcompleteness levels are indicated.

\begin{table}[t]
\begin{center}
\begin{tabular}{|c|c|c|c|c|}
\hline Number of bases & \multicolumn{2}{ |c| }{\textit{Architecture 1}} &  \multicolumn{2}{ |c| }{\textit{Architecture 2}} \\
\cline{2-3}
\cline{4-5}
 & \textit{Liberty}  dataset & \textit{Notredame}  dataset & \textit{Liberty}  dataset & \textit{Notredame}  dataset
 \\ \hline \hline
4097 &   0.8096 & 0.6126 & 0.8175 & 0.6634 \\

5000 & 0.8264 & 0.6669 & 0.8300 & 7605 \\

6000 & 0.8373 & 0.7961 & 0.8506 & 0.8818 \\

7000 & \textbf{0.8423} & \textbf{0.8357} & \textbf{0.8578} & \textbf{0.8970} \\
\hline
\end{tabular}
\end{center}
\caption{Maximum validation accuracy recorded.}
\label{maxumumvalidation}
\end{table} 

\begin{table}[t]
\begin{center}
\begin{tabular}{|l|c|c|}
\hline Training & \textit{Liberty} & \textit{Notredame} \\
Test & \textit{Notredame} & \textit{Liberty} \\ \hline \hline
nSIFT + L2  & 22.53  & 29.84\\
nSIFT  + NNet & 14.35 & 20.44 \\
Trzcinski \textit{et al} \cite{trzcinski2012learning} & 14.15 & 18.05 \\
Brown \textit{et al} \cite{brown2011discriminative} & - & 16.85 \\
Simonyan \textit{et al} \cite{simonyan2014learning} & 9.88 & 16.56 \\
Han \textit{et al} \cite{han2015matchnet} & 3.87 & 6.90 \\
Zagoruyko and  Komodakis  \cite{zagoruyko2015learning}  & 4.56 & 2.01\\
Our method ( k = 4097, \textit{Architecture 1}) & 13.52 & 16.17\\
Our method ( k = 5000, \textit{Architecture 1}) & 7.78 & 10.39\\
Our method ( k = 6000,\textit{Architecture 1}) & 4.26 & 5.69\\
Our method ( k = 7000, \textit{Architecture 1}) & 3.59 & 4.36\\
Our method ( k = 4097, \textit{Architecture 2}) & 9.69 & 12.11\\
Our method ( k = 5000, \textit{Architecture 2}) & 6.45 & 7.12\\
Our method ( k = 6000, \textit{Architecture 2}) & 3.19 & 4.13\\
Our method ( k = 7000, \textit{Architecture 2}) & \textbf{2.67} & \textbf{1.89}\\
\hline
\end{tabular}
\end{center}
\caption{Error@95\% on UBC dataset.} 
\label{resultstable}
\end{table} 

To compare our results with the previous approaches we used the evaluations carried out by the previous comparisons in the work by Han \textit{et al.} \cite{han2015matchnet} and in the work by Zagoruyko and Komodakis \cite{zagoruyko2015learning}. The obtain results are available in Table \ref{resultstable}. The settings of the baseline methods used for the comparison are described in the following paragraph.

To compare SIFT feature's \cite{lowe2004distinctive} patch matching ability we used the same experimental results which have been obtained in \cite{han2015matchnet}. They have used the SIFT features implemented by VL Feat \cite{vedaldi08vlfeat} with the bin size of 16. Normalized SIFT (nSIFT) has been obtained by  scaling the original SIFT feature such that its L2 norm is 1. In Table \ref{resultstable} the patch comparison based on nSIFT and L2 distance is denoted by \textit{nSIFT + L2}.  \textit{nSIFT + NNet} stands for a framework where nSIFT features have been used as the input for a neural network with the 2 fully connected with 512 neurons in each. 150k iterations have been used to train this network. To compare the work by Trzcinski \textit{et al} \cite{trzcinski2012learning}, Brown \textit{et al} \cite{brown2011discriminative}, Han \textit{et al} \cite{han2015matchnet} and Zagoruyko and  Komodakis \cite{zagoruyko2015learning} we used their results which have been obtained under the best configuration.

When the results are compared it can be identified that our setting of using 7000 bases with the \textit{Architecture 2} has yielded better results  when compared with the results obtained in the state-of-the-art methods \cite{han2015matchnet,zagoruyko2015learning} for both datasets. Moreover when considering \textit{Liberty} dataset, our setting with 6000 bases has outperformed the method by Zagoruyko and Komodakis \cite{zagoruyko2015learning} for the setting of 6000 bases with the \textit{Architecture 1} and \textit{Architecture 2} and for the setting with 7000 bases with \textit{Architecture 1}. In addition for the  \textit{Liberty} dataset our setting with 6000 bases with \textit{Architecture 2} and 7000 bases with \textit{Architecture 1} has outformed the method by Han \textit{et al.} \cite{han2015matchnet}. When considering the \textit{Notredame} dataset our method has outperformed the method by Han \textit{et al.} \cite{han2015matchnet} for the setting of 6000 bases with \textit{Architecture 1} and \textit{Architecture 2} and 7000 bases with \textit{Architecture 1} as well. From these results it can be observed that increasing overcompleteness has yeilded better results.

\section{Conclusion}
\label{conclusion}
In this paper we have introduced and evaluated a novel patch descriptor by encoding patches using a dictionary and a neural network to compare the generated descriptor. To this  end we studied two main neural network architectures and our model has outperformed the state-of-the-art accuracy on the standard UBC patch dataset. Among the architectures we suggest the architecture with one hidden layer which is shown to yield  the best results. In contrast to previous patch descriptors which use visual information of the patch images, we use coefficients of sparse coding representation of the patch images capturing  the statistical information in  the patches . The use of  neural network based patch comparison enables the technique to effectively  match  the obtained coefficient by  capturing the non linear relationships across them. Furthermore we identified that increasing overcompleteness can result in better accuracy due to the fact that overcomplete representation is better able to capture  the structures and patterns that are present in the original input image.

\end{document}